\documentclass{article}

\PassOptionsToPackage{numbers,sort&compress}{natbib}
\usepackage[final,sglblindworkshop]{neurips_2026}
\workshoptitle{Second Workshop on MLxOR: Mathematical Foundations and Operational Integration of Machine Learning for Uncertainty-Aware Decision-Making}

\usepackage[utf8]{inputenc}
\usepackage[T1]{fontenc}
\usepackage{amsmath,amssymb}
\usepackage{booktabs}
\usepackage[font=small,labelfont=bf]{caption}
\usepackage{float}
\usepackage{graphicx}
\usepackage{microtype}
\usepackage{xcolor}
\usepackage{hyperref}
\usepackage{url}
\hypersetup{hidelinks}

\newcommand{\method}{OutageDiT}
\newcommand{\R}{\mathbb{R}}

\title{\method: A Foundation Model for Power Outage Forecasting and Scenario Generation}

\author{%
  Yunqin Zhu$^1$ \quad Feng Qiu$^2$ \quad Yao Xie$^1$\\
  $^1$H. Milton Stewart School of Industrial and Systems Engineering,\\
  Georgia Institute of Technology\\
  $^2$Argonne National Laboratory\\
  \texttt{yao.xie@isye.gatech.edu}
}

\begin{document}
\maketitle

\begin{abstract}
Power-outage planning requires scenarios before an event occurs. These scenarios must represent uncertainty in magnitude, timing, and duration
while preserving temporal dependence. However, severe events are rare, and data
from any single region contain few examples of extreme outage and restoration
patterns. To address this challenge, we introduce \method{}, a foundation model for generating seven-day outage trajectories at 15-minute
resolution, trained on outage and weather records across the United States. Specifically, a condition encoder processes historical context and known future covariates once per forecast, and a shallow flow-matching decoder reuses the resulting horizon-aligned states to sample complete outage trajectories. This enables point forecasting, uncertainty quantification, and conditional scenario generation within one deep generative model.
Across outage forecasting benchmarks, \method{} improves forecast accuracy and
scenario quality over strong baselines and supports zero-shot transfer to
held-out regions. Together, these results position conditional outage scenario generation
as a bridge from outage forecasting to operational planning under uncertainty.
\end{abstract}

\section{Introduction}

Power-outage response decisions are made before an event's magnitude and
duration are fully known. Examples include crew staging and resource allocation, which require probabilistic forecasts of how outage counts may evolve over the planning horizon under uncertainty.
Scenario-based planning further requires representative uncertainty realizations
\citep{bertsimas2020prescriptive,kaut2007scenario}. For outage response, these are complete trajectories that preserve
temporal dependence and plausible patterns of escalation, peak severity, and
restoration. We formulate this task as generating samples from the conditional distribution
\begin{equation}
  p\,\bigl(
    \underbrace{y_{0:T}}_{\text{future outages}}\;\bigm|\;
    \underbrace{h_{-L:0}}_{\text{outage and customer history}},
    \underbrace{w_{-L:T}}_{\text{weather covariates}},
    \underbrace{a_{-L:T}}_{\text{calendar covariates}}
  \bigr),
  \label{eq:forecast-law}
\end{equation}
where $L$ and $T$ denote the numbers of 15-minute time steps in the 14-day
history window and seven-day forecast horizon, respectively. Outage trajectories
depend on weather as well as local infrastructure,
vegetation, and grid vulnerability
\citep{guikema2014,he2017,cerrai2019,watson2022,taylor2023,zhu2025gridresilience}.
Yet severe events are rare within any single region, leaving limited examples of
how outages escalate and recover under different conditions. Cross-region
pretraining can broaden this coverage by learning shared outage and restoration
patterns, while conditioning on local history, weather, and calendar covariates
retains regional context. These considerations motivate a foundation model that
combines shared patterns with local context to support transfer to regions with
little or no local training data.

We develop \method{}, a generative foundation model for power outage
trajectories, pretrained on outage and weather records across the United States.
To incorporate local context, a condition encoder maps 14 days of history and
covariates available at forecast time to states aligned with the forecast horizon.
A shallow decoder
based on the Diffusion Transformer (DiT) architecture \citep{dit2023,sit2024}
then reuses these states to capture temporal dependence in future outages.
Trained with flow matching \citep{flowmatching2023}, the decoder generates
trajectories through ODE integration. The resulting trajectories are decoded into
nonnegative integer outage counts using an analytic count representation.
Across national seven-day forecasting and held-out Michigan benchmarks,
\method{} improves point and probabilistic forecast accuracy over supervised and
fine-tuned foundation-model baselines, better represents temporal dependence and
forecast uncertainty in generated scenarios, and transfers directly to a state excluded from training.

\section{Method}

\paragraph{Count representation.}
Outage and tracked-customer counts are nonnegative integers with a wide dynamic
range (up to $10^7$ in our dataset). We represent overall scale and digit-level
detail separately, with an analytic inverse that recovers integer counts.
Direct decimal encoding, however, is discontinuous at carry boundaries: an
increment of one can change several digits. For a count $n$, let
$(d_6,\ldots,d_0)$ be its zero-padded decimal digits, with $d_0$ the units digit.
We reflect each lower digit when the next higher digit is odd:
\[
  r_6=d_6,\qquad
  r_j=d_j+(d_{j+1}\bmod 2)(9-2d_j),
  \quad j=0,\ldots,5.
\]
These carry-aware coordinates keep consecutive counts adjacent across carries.
We also append a normalized log-magnitude coordinate to provide a monotone
summary of scale and determine which decimal positions are valid during decoding. The resulting
representation is
\begin{equation}
  g(n)=\left[2\log_{10}(1+n)/7-1,
  (2r_6-9)/10,\ldots,(2r_0-9)/10\right]
  \in\R^8.
  \label{eq:count-transform}
\end{equation}

\paragraph{Condition encoder.}
Recent outages describe the current state, while longer history provides context
for how outages respond to weather. We therefore use 14 days of transformed
quarter-hour outage and tracked-customer counts with weather and calendar
covariates, and represent the most recent day in a separate stream.
Four consecutive observations
are patched into an hourly token. The condition encoder applies transformer
blocks \citep{vaswani2017attention} to 336 history tokens, an overlapping
24-token recent count stream, and 168 future tokens formed from weather and calendar variables
known over the seven-day horizon. A shared signed hourly time axis aligns these
streams through rotary position embeddings (RoPE) \citep{su2024roformer}.
History serves as a read-only key-value bank, allowing recent and future tokens
to attend jointly to historical context without updating the history tokens.
The final encoder block returns 168 horizon-aligned
conditioning states $c=(c_0,\ldots,c_{167})$.

\paragraph{Flow decoder.}
We apply flow matching \citep{flowmatching2023} to generate complete
trajectories and model uncertainty jointly across the forecast horizon.
To apply this continuous formulation to discrete digit coordinates, we add
independent noise $\epsilon_j\sim\mathcal{U}(-0.05,0.05)$ to each future digit
coordinate at every forecast time step without crossing quantization-bin
boundaries. Let $\tilde y$ denote the trajectory obtained by applying $g$ to each
raw count in $y=y_{0:T}$ and adding this noise. With $z\sim\mathcal{N}(0,I)$,
we sample $\tau\sim\mathcal{U}(0,1)$, construct
$\tilde y_\tau=(1-\tau)\tilde y+\tau z$, and train the decoder to predict velocity $z-\tilde y$.
Since the conditioning states remain fixed while the noisy trajectory changes,
a shallow decoder reuses $c$ to predict this velocity. It first patchifies
$\tilde y_\tau\in\R^{T\times8}$ into 168 hourly tokens.
Each layer then jointly attends to the conditioning states $c$ and noisy
future tokens using a single softmax, combining local context with temporal
interactions across the forecast horizon. Within each layer, flow time $\tau$
modulates the attention and feed-forward residuals through adaptive layer
normalization (adaLN) \cite{dit2023}. Finally, the output head maps each hourly token to
four quarter-hour velocity predictions. We also use an auxiliary head $f_\theta$ to predict $\log(1+y)$ from $c$ at 15-minute resolution,
providing direct forecasting supervision to the condition encoder. The
full objective is
\begin{equation}
  \mathcal{L}=\mathbb{E}\,\Bigl[
  \underbrace{\frac{1}{8T}
  \|v_\theta(\tilde y_\tau,\tau,c)-(z-\tilde y)\|_F^2}_{\text{flow matching loss}}
  +\lambda_{\mathrm{aux}}
  \underbrace{\frac{1}{T}
  \|f_\theta(c)-\log(1+y)\|_F^2}_{\text{auxiliary loss}}
  \Bigr],
  \label{eq:fm-loss}
\end{equation}
where $\lambda_{\mathrm{aux}}$ is the auxiliary-loss weight.
At inference, the condition encoder runs once, and its layer-wise
key-value projections are cached across Monte Carlo samples and ODE solver
steps. Sampling starts from $\tilde y_1=z$ and integrates the flow ODE
from $\tau=1$ to $\tau=0$, updating only the noisy future tokens and
their projections. After integration, $g^{-1}$ maps the transformed
coordinates back to integer counts. Appendix~\ref{app:architecture-optimization} provides the complete architecture and optimization details.

\section{Experiments}

\paragraph{Experimental setup.}
Training uses U.S. county-level windows outside Michigan ending before 2025.
Validation covers January--February 2025 and test covers March--May 2025. Each
evaluation split contains 30 normal windows and 30 event windows drawn from ten distinct
events. An event must exceed both 800 outages and 4\% of tracked customers for
at least six hours. Inputs combine historical outage and customer counts, calendar variables, and
84 hourly meteorological covariates derived from NOAA/NCEP High-Resolution
Rapid Refresh (HRRR) fields\footnote{\url{https://emc.ncep.noaa.gov/emc/pages/numerical_forecast_systems/hrrr.php}};
weather is observed through the forecast horizon. All supervised models use
the same county/origin sampling distribution, and validation MSE selects
checkpoints. We report MSE and weighted quantile loss (WQL) in
$\log_{10}(1+y)$ space; lower values are better. Appendices~\ref{app:data},
\ref{app:architecture-optimization}, \ref{app:national-metrics}, and
\ref{app:michigan-metrics} provide the data, training, and evaluation details.

\begin{figure}[H]
  \centering
  \includegraphics[width=0.98\linewidth]{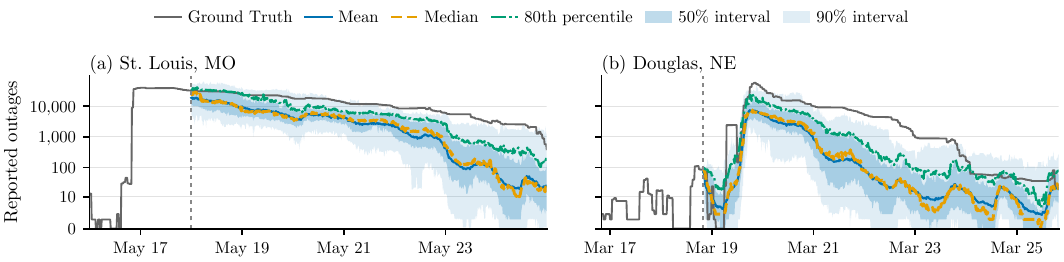}
  \caption{Outage forecasts for (a) restoration in St. Louis, MO and
  (b) event onset in Douglas, NE. Dashed vertical lines mark forecast origins;
  bands show central 50\% and 90\% intervals. Values use the
  \(\log_{10}(1+y)\) scale; dates are in 2025.}
  \label{fig:trajectory}
\end{figure}

\paragraph{National forecasting.}
Table~\ref{tab:national-ablations}(a) compares \method{} with task-trained
probabilistic baselines \citep{deepar2020,tft2021,csdi2021} and fine-tuned
time-series foundation models \citep{timesfm2024,chronos2_2025}. \method{}
performs best on every split and metric. The advantage holds on normal and
event windows, so the aggregate gain is not driven by a single regime.
Table~\ref{tab:national-ablations}(b) shows that every conditioning branch
contributes. Future conditioning has the largest effect, while the history and
recent ablations confirm that both temporal scales remain useful. Digit
coordinates and the auxiliary log-count loss provide the largest gains among
the remaining components; uniform digit noise yields a smaller but consistent
improvement. Figure~\ref{fig:trajectory} illustrates complementary high-impact dynamics:
the forecast follows restoration and secondary fluctuations in St. Louis, MO,
while the Douglas, NE forecast captures event onset one day before the
ground-truth peak.

\begin{table*}[t]
  \caption{National seven-day forecasting performance and ablations. All metrics evaluate
  outage count in $\log_{10}(1+y)$ space. Ablations use the same training
  budget and evaluation protocol.}
  \label{tab:national-ablations}
  \centering
  \footnotesize
  \begin{minipage}[t]{0.60\textwidth}
    \centering
    \textbf{(a) Overall performance}\par\vspace{3pt}
    \setlength{\tabcolsep}{4.0pt}
    \begin{tabular}{lrrrrrr}
      \toprule
      & \multicolumn{3}{c}{MSE} & \multicolumn{3}{c}{WQL} \\
      \cmidrule(lr){2-4}\cmidrule(lr){5-7}
      Model & All & Normal & Event & All & Normal & Event \\
      \midrule
      DeepAR & 2.356 & 0.317 & 4.395 & 0.664 & 0.874 & 0.639 \\
      TFT & 1.853 & 0.207 & 3.499 & 0.523 & 0.589 & 0.515 \\
      CSDI & 1.445 & 0.358 & 2.533 & 0.489 & 0.906 & 0.440 \\
      TimesFM 2.5 & 1.233 & 0.209 & 2.257 & 0.444 & 0.918 & 0.388 \\
      Chronos-2 & 1.041 & 0.190 & 1.892 & 0.361 & 0.550 & 0.339 \\
      \method{} & \textbf{0.857} & \textbf{0.162} & \textbf{1.552}
        & \textbf{0.328} & \textbf{0.508} & \textbf{0.306} \\
      \bottomrule
    \end{tabular}
  \end{minipage}\hfill
  \begin{minipage}[t]{0.38\textwidth}
    \centering
    \textbf{(b) Ablations of OutageDiT}\par\vspace{3pt}
    \setlength{\tabcolsep}{4.0pt}
    \begin{tabular}{lccc}
      \toprule
      Variant & MSE & WQL & VS \\
      \midrule
      -- & \textbf{0.857} & \textbf{0.328} & \textbf{0.151} \\
      w/o history & 0.929 & 0.357 & 0.163 \\
      w/o recent & 0.962 & 0.356 & 0.156 \\
      w/o future & 1.346 & 0.424 & 0.175 \\
      \midrule
      w/o digits & 0.903 & 0.349 & 0.158 \\
      w/o dequant. & 0.869 & 0.331 & 0.152 \\
      w/o aux. loss & 0.912 & 0.345 & 0.154 \\
      \bottomrule
    \end{tabular}
  \end{minipage}
\end{table*}

\begin{figure}[H]
  \centering
  \includegraphics[width=0.88\linewidth]{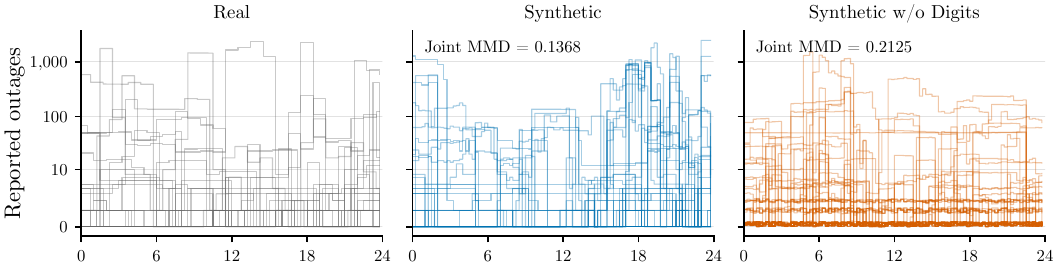}
  \caption{Random one-day outage trajectories. Each panel shows 64
  trajectories over a 24-hour forecast horizon. Joint MMD is the empirical
  RBF-kernel discrepancy over complete trajectories, with bandwidth set to the
  median pairwise RMS distance among real trajectories.}
  \label{fig:trajectory-distribution}
\end{figure}

\paragraph{Scenario generation.}
For models returning complete trajectories, we report the variogram score (VS)
and the coverage and width of the central 90\% prediction interval. VS is sensitive
to temporal dependence; coverage and width assess marginal calibration and
sharpness. We also compare with a generative time-series foundation model
\citep{sundial2025}. Table~\ref{tab:simulation-transfer}(a) shows that \method{} has the
lowest VS and highest coverage with interval width comparable to competing
methods. Lower coverage on event windows identifies extreme-event calibration
as the primary limitation. Figure~\ref{fig:trajectory-distribution} further
shows that digit coordinates bring synthetic one-day trajectories closer to
the real joint distribution, consistent with their gains beyond pointwise
forecast error.

\paragraph{Geographic transfer.}
Michigan provides a geographically held-out evaluation: \method{} is trained
without Michigan data and applied directly to the state. The INFORMS DMDA
Workshop Data Challenge reports average 24/48-hour
RMSE\footnote{\url{https://sites.google.com/view/dmdaworkshop2025/data-challenge}};
SARIMAX is its best-performing forecasting method \citep{ye2025sarimax}. The
IISE Energy Analytics Challenge reports normal and tail RMSE, nonzero F1, and
95\% interval Winkler
score\footnote{\url{https://www.iise.org/Details.aspx?id=54646}}. We use the
official scores of its winner, the Regime-Switching Time Series Foundation
Model (RSFM). Table~\ref{tab:simulation-transfer}(b) shows that \method{}
leads every reported transfer metric without Michigan training. The gains span
average, normal, and tail errors, nonzero detection, and interval quality,
supporting transfer across outage regimes and forecast summaries.

\begin{table}[t]
  \caption{Scenario quality and zero-shot Michigan transfer. National
  coverage targets 0.90; Michigan metrics follow the DMDA and IISE challenge
  protocols.}
  \label{tab:simulation-transfer}
  \centering
  \footnotesize
  \vspace{3pt}
  \begin{minipage}[t]{0.38\linewidth}
    \centering
    \textbf{(a) National seven-day forecasting}\par\vspace{2pt}
    \setlength{\tabcolsep}{4.0pt}
    \begin{tabular}{lccc}
      \toprule
      Model & VS & Coverage & Width \tabularnewline
      \midrule
      Sundial & 0.207 & 0.650 & 1.471 \tabularnewline
      DeepAR & 0.354 & 0.622 & 1.161 \tabularnewline
      CSDI & 0.234 & 0.607 & 1.877 \tabularnewline
      \method{} & \textbf{0.151} & \textbf{0.839} & 1.483 \tabularnewline
      \bottomrule
    \end{tabular}
  \end{minipage}\hfill
  \begin{minipage}[t]{0.60\linewidth}
    \centering
    \textbf{(b) Michigan competition benchmarks}\par\vspace{2pt}
    \setlength{\tabcolsep}{4pt}
    \begin{tabular}{lccccc}
      \toprule
      & \multicolumn{1}{c}{Avg.} & \multicolumn{1}{c}{Normal} & \multicolumn{1}{c}{Tail} & \multicolumn{1}{c}{Nonzero} & \tabularnewline
      Model & \multicolumn{1}{c}{RMSE} & \multicolumn{1}{c}{RMSE} & \multicolumn{1}{c}{RMSE} & \multicolumn{1}{c}{F1} & \multicolumn{1}{c}{Winkler} \tabularnewline
      \midrule
      SARIMAX & 177.24 & 59.82 & 264.28 & 0.577 & -- \tabularnewline
      RSFM & 175.59 & 12.89 & 286.61 & 0.649 & 2615 \tabularnewline
      \method{} & \textbf{164.21} & \textbf{11.82} & \textbf{257.58} & \textbf{0.715} & \textbf{2479} \tabularnewline
      \bottomrule
    \end{tabular}
  \end{minipage}
\end{table}

\section{Conclusion}

This work introduces \method{}, a generative foundation model for conditional
seven-day outage trajectories learned from county-level records across the
United States. On national benchmarks, \method{} improves point and
probabilistic forecast accuracy and better preserves scenario-level temporal
dependence; ablations attribute these gains to the full conditioning structure
and count representation. On Michigan, which is excluded from training, \method{} outperforms the
reported DMDA and IISE competition baselines across all metrics, demonstrating
zero-shot geographic transfer. Extreme-event undercoverage remains the
primary limitation. Future work should evaluate these scenarios within
crew-staging and resource-allocation decisions, account for weather-forecast
error, and extend nominal forecasts to worst-case planning under distribution
shift \citep{cheng2026generative,zhu2024dro,cheng2025worstcase}.

\section*{Acknowledgment}

The project is partially supported by the U.S. Department of Energy Advanced Grid Modeling Program under Grant DE-OE 0000875.
We thank the NVIDIA Academic Grant Program for providing H100 GPU computing resources through NVIDIA Brev.

\bibliographystyle{plainnat}
\bibliography{references}

@article{guikema2014,
  author  = {Guikema, Seth David and Nateghi, Roshanak and Quiring, Steven M. and Staid, Andrea and Reilly, Allison C. and Gao, Michael},
  title   = {Predicting Hurricane Power Outages to Support Storm Response Planning},
  journal = {IEEE Access},
  volume  = {2},
  pages   = {1364--1373},
  year    = {2014},
}

@article{he2017,
  author  = {He, Jichao and Wanik, David W. and Hartman, Brian M. and Anagnostou, Emmanouil N. and Astitha, Marina and Frediani, Maria E. B.},
  title   = {Nonparametric Tree-Based Predictive Modeling of Storm Outages on an Electric Distribution Network},
  journal = {Risk Analysis},
  volume  = {37},
  number  = {3},
  pages   = {441--458},
  year    = {2017},
}

@article{cerrai2019,
  author  = {Cerrai, Diego and Wanik, David W. and Bhuiyan, Md Abul Ehsan and Zhang, Xinxuan and Yang, Jaemo and Frediani, Maria E. B. and Anagnostou, Emmanouil N.},
  title   = {Predicting Storm Outages Through New Representations of Weather and Vegetation},
  journal = {IEEE Access},
  volume  = {7},
  pages   = {29639--29654},
  year    = {2019},
}

@article{watson2022,
  author  = {Watson, Peter L. and Spaulding, Aaron and Koukoula, Marika and Anagnostou, Emmanouil},
  title   = {Improved quantitative prediction of power outages caused by extreme weather events},
  journal = {Weather and Climate Extremes},
  volume  = {37},
  pages   = {100487},
  year    = {2022},
}

@article{taylor2023,
  author  = {Taylor, William O. and Cerrai, Diego and Wanik, David and Koukoula, Marika and Anagnostou, Emmanouil N.},
  title   = {Community power outage prediction modeling for the {Eastern United States}},
  journal = {Energy Reports},
  volume  = {10},
  pages   = {4148--4169},
  year    = {2023},
}

@inproceedings{timesfm2024,
  author    = {Das, Abhimanyu and Kong, Weihao and Sen, Rajat and Zhou, Yichen},
  title     = {A decoder-only foundation model for time-series forecasting},
  booktitle = {Proceedings of the 41st International Conference on Machine Learning},
  series    = {Proceedings of Machine Learning Research},
  volume    = {235},
  pages     = {10148--10167},
  year      = {2024},
  publisher = {PMLR}
}

@article{chronos2_2025,
  author  = {Ansari, Abdul Fatir and Shchur, Oleksandr and K{\"u}ken, Jaris and Auer, Andreas and Han, Boran and Mercado, Pedro and Rangapuram, Syama Sundar and Shen, Huibin and Stella, Lorenzo and Zhang, Xiyuan and Goswami, Mononito and Kapoor, Shubham and Maddix, Danielle C. and Guerron, Pablo and Hu, Tony and Yin, Junming and Erickson, Nick and Desai, Prateek Mutalik and Wang, Hao and Rangwala, Huzefa and Karypis, George and Wang, Yuyang and Bohlke-Schneider, Michael},
  title   = {{Chronos-2}: From Univariate to Universal Forecasting},
  journal = {arXiv preprint arXiv:2510.15821},
  year    = {2025},
}

@inproceedings{sundial2025,
  author    = {Liu, Yong and Qin, Guo and Shi, Zhiyuan and Chen, Zhi and Yang, Caiyin and Huang, Xiangdong and Wang, Jianmin and Long, Mingsheng},
  title     = {Sundial: A Family of Highly Capable Time Series Foundation Models},
  booktitle = {Proceedings of the 42nd International Conference on Machine Learning},
  series    = {Proceedings of Machine Learning Research},
  volume    = {267},
  pages     = {39295--39317},
  year      = {2025},
  publisher = {PMLR}
}

@article{deepar2020,
  author  = {Salinas, David and Flunkert, Valentin and Gasthaus, Jan and Januschowski, Tim},
  title   = {{DeepAR}: Probabilistic forecasting with autoregressive recurrent networks},
  journal = {International Journal of Forecasting},
  volume  = {36},
  number  = {3},
  pages   = {1181--1191},
  year    = {2020},
}

@article{tft2021,
  author  = {Lim, Bryan and Ar{\i}k, Sercan {\"O}. and Loeff, Nicolas and Pfister, Tomas},
  title   = {{Temporal Fusion Transformers} for interpretable multi-horizon time series forecasting},
  journal = {International Journal of Forecasting},
  volume  = {37},
  number  = {4},
  pages   = {1748--1764},
  year    = {2021},
}

@inproceedings{csdi2021,
  author    = {Tashiro, Yusuke and Song, Jiaming and Song, Yang and Ermon, Stefano},
  title     = {{CSDI}: Conditional Score-based Diffusion Models for Probabilistic Time Series Imputation},
  booktitle = {Advances in Neural Information Processing Systems},
  volume    = {34},
  pages     = {24804--24816},
  year      = {2021}
}

@inproceedings{vaswani2017attention,
  author    = {Vaswani, Ashish and Shazeer, Noam and Parmar, Niki and Uszkoreit, Jakob and Jones, Llion and Gomez, Aidan N. and Kaiser, {\L}ukasz and Polosukhin, Illia},
  title     = {Attention Is All You Need},
  booktitle = {Advances in Neural Information Processing Systems},
  volume    = {30},
  year      = {2017}
}

@article{su2024roformer,
  author  = {Su, Jianlin and Ahmed, Murtadha and Lu, Yu and Pan, Shengfeng and Bo, Wen and Liu, Yunfeng},
  title   = {{RoFormer}: Enhanced transformer with Rotary Position Embedding},
  journal = {Neurocomputing},
  volume  = {568},
  pages   = {127063},
  year    = {2024},
}

@inproceedings{flowmatching2023,
  author    = {Lipman, Yaron and Chen, Ricky T. Q. and Ben-Hamu, Heli and Nickel, Maximilian and Le, Matt},
  title     = {Flow Matching for Generative Modeling},
  booktitle = {International Conference on Learning Representations},
  year      = {2023},
  url       = {https://openreview.net/forum?id=PqvMRDCJT9t}
}

@inproceedings{dit2023,
  author    = {Peebles, William and Xie, Saining},
  title     = {Scalable Diffusion Models with Transformers},
  booktitle = {2023 IEEE/CVF International Conference on Computer Vision (ICCV)},
  pages     = {4172--4182},
  year      = {2023},
}

@inproceedings{sit2024,
  author    = {Ma, Nanye and Goldstein, Mark and Albergo, Michael S. and Boffi, Nicholas M. and Vanden-Eijnden, Eric and Xie, Saining},
  title     = {{SiT}: Exploring Flow and Diffusion-Based Generative Models with Scalable Interpolant Transformers},
  booktitle = {Computer Vision -- ECCV 2024},
  series    = {Lecture Notes in Computer Science},
  volume    = {15135},
  pages     = {23--40},
  year      = {2024},
  publisher = {Springer},
}

@article{scheuerer2015variogram,
  author  = {Scheuerer, Michael and Hamill, Thomas M.},
  title   = {Variogram-Based Proper Scoring Rules for Probabilistic Forecasts of Multivariate Quantities},
  journal = {Monthly Weather Review},
  volume  = {143},
  number  = {4},
  pages   = {1321--1334},
  year    = {2015},
}

@article{ye2025sarimax,
  author  = {Ye, Haoran and Sun, Qiuzhuang and Yang, Yang},
  title   = {{SARIMAX}-Based Power Outage Prediction During Extreme Weather Events},
  journal = {arXiv preprint arXiv:2511.01017},
  year    = {2025},
}

@article{zhu2025gridresilience,
  author  = {Zhu, Shixiang and Yao, Rui and Xie, Yao and Qiu, Feng and Qiu, Yueming and Wu, Xuan},
  title   = {Quantifying Grid Resilience Against Extreme Weather Using Large-Scale Customer Power Outage Data},
  journal = {INFORMS Journal on Data Science},
  volume  = {5},
  number  = {2},
  pages   = {102--118},
  year    = {2026},
}

@article{bertsimas2020prescriptive,
  author  = {Bertsimas, Dimitris and Kallus, Nathan},
  title   = {From Predictive to Prescriptive Analytics},
  journal = {Management Science},
  volume  = {66},
  number  = {3},
  pages   = {1025--1044},
  year    = {2020},
}

@article{kaut2007scenario,
  author  = {Kaut, Michal and Wallace, Stein W.},
  title   = {Evaluation of Scenario-Generation Methods for Stochastic Programming},
  journal = {Pacific Journal of Optimization},
  volume  = {3},
  number  = {2},
  pages   = {257--271},
  year    = {2007}
}

@inproceedings{zhu2024dro,
  author  = {Zhu, Linglingzhi and Zhu, Yunqin and Xie, Yao},
  title   = {Distributionally Robust Optimization via Iterative Algorithms in Continuous Probability Spaces},
  booktitle = {NeurIPS 2025 ML$\times$OR Workshop},
  year    = {2025},
  url     = {https://openreview.net/forum?id=xDRGZLcuN8}
}

@article{cheng2025worstcase,
  author  = {Cheng, Xiuyuan and Xie, Yao and Zhu, Linglingzhi and Zhu, Yunqin},
  title   = {Worst-case generation via minimax optimization in {Wasserstein} space},
  journal = {arXiv preprint arXiv:2512.08176},
  year    = {2025},
}

@article{cheng2026generative,
  author  = {Cheng, Xiuyuan and Zhu, Yunqin and Xie, Yao},
  title   = {Generative models for decision-making under distributional shift},
  journal = {arXiv preprint arXiv:2604.04342},
  year    = {2026},
}

\clearpage
\appendix
\section{Data and Evaluation Protocol}
\label{app:data}

\subsection{Data and Inputs}

The training dataset contains quarter-hour outage and tracked-customer counts
from U.S. counties outside Michigan. Michigan forms a geographically held-out
transfer benchmark. Outage
records span 2018--2024; complete weather covariates are available
from January 2022. Each example contains 14 days of context and a seven-day
target, corresponding to 1,344 and 672 quarter-hour observations. The model
receives outage and customer-count history, 84 weather variables, and eight
calendar variables. Its conditioning path includes observed weather and
calendar values over the seven-day forecast horizon. The hourly weather
variables are derived from NOAA/NCEP HRRR fields supplied on a county-indexed
grid and span temperature, wind, pressure, moisture, cloud, precipitation,
convection, and radar diagnostics. Each weather feature is standardized using
pre-2025 training data and aligned to its four corresponding quarter-hour
observations.

\subsection{Training and Evaluation Splits}

Training windows follow the natural county-date distribution: each draw first
selects a county and then a valid forecast origin in that county. The complete
14-day history stream is retained
independently for half of the training examples, while the overlapping 24-hour
recent stream is always retained. Weather conditioning is augmented with
probabilities 0.80 for full past and future weather, 0.10 for past weather with
future weather masked, and 0.10 for all weather masked.

The national validation set contains 30 event and 30 normal windows with
origins from January 15 through February 20, 2025; the test set contains the
same composition with origins from March 16 through May 23, 2025. The 30 event
windows in each split are formed from ten distinct events with three forecast
origins per event. An event must exceed both 800 outages and 4\% of tracked
customers continuously for at least six hours. Normal windows are drawn from
counties outside the selected events. Every national score gives equal weight
to each forecast case.

\section{Architecture and Optimization}
\label{app:architecture-optimization}

Four consecutive quarter-hour observations are patchified into one hourly
token, yielding 336 history, 24 recent, and 168 future tokens. The condition
encoder has four transformer blocks; the flow decoder has two. Both use width
1,024, 16 attention heads with per-head dimension $d_h=64$, and a feed-forward
expansion factor of four. The complete model has 183.46 million trainable
parameters. Outage count is the
prediction target, and tracked-customer count is a historical covariate.

\subsection{Tokenization and Position}
We use the outage and customer history $h$, weather covariates $w$, and
calendar covariates $a$ from Eq.~\eqref{eq:forecast-law}. All slices exclude the
right endpoint.
Token matrices store tokens in columns; $\operatorname{LN}$ denotes layer
normalization over features, and biases are broadcast across tokens. Layer
superscripts distinguish token states from raw time-indexed inputs.
For a stream $u$ at 15-minute resolution, locally indexed from zero, each modality $m$
uses the two-layer patch map
\begin{equation}
  \operatorname{Patch}_m(u)_i
  =W_{m,2}\operatorname{SiLU}\!\left(
    W_{m,1}\operatorname{vec}(u_{4i},\ldots,u_{4i+3})+b_{m,1}
  \right)+b_{m,2}.
  \label{eq:patch-map}
\end{equation}
Here $\operatorname{vec}$ stacks four observations into a column vector.
For count inputs, the transform $g$ in Eq.~\eqref{eq:count-transform} is applied
to each count, and the outage and customer coordinates are concatenated.
The three condition streams are
\begin{equation}
  \begin{aligned}
    h^{(0)}&=\operatorname{LN}_h\!\left(
      \operatorname{Patch}_n(g(h_{-L:0}))+\operatorname{Patch}_w(w_{-L:0})+\operatorname{Patch}_a(a_{-L:0})\right),\\
    r^{(0)}&=\operatorname{LN}_r\!\left(
      \operatorname{Patch}_n(g(h_{-96:0}))+\operatorname{Patch}_a(a_{-96:0})\right),\\
    f^{(0)}&=\operatorname{LN}_f\!\left(
      \operatorname{Patch}_w(w_{0:T})+\operatorname{Patch}_a(a_{0:T})\right),
  \end{aligned}
  \label{eq:condition-tokens}
\end{equation}
where $n,w,a$ denote count, weather, and calendar modalities. Missing count or
weather patches are zeroed before summation. For zero-based hourly index $i$,
the signed positions are $\mathrm{pos}_i^{h}=i+\tfrac12-336$,
$\mathrm{pos}_i^{r}=i+\tfrac12-24$, and $\mathrm{pos}_i^{f}=i+\tfrac12$.
RoPE rotates a query/key vector $x$ at its signed position $\mathrm{pos}$:
\begin{equation}
  \begin{bmatrix}
    \operatorname{RoPE}(x,\mathrm{pos})_{2k}\\
    \operatorname{RoPE}(x,\mathrm{pos})_{2k+1}
  \end{bmatrix}
  =
  \begin{bmatrix}
    \cos(\mathrm{pos}\,\omega_k)&-\sin(\mathrm{pos}\,\omega_k)\\
    \sin(\mathrm{pos}\,\omega_k)& \cos(\mathrm{pos}\,\omega_k)
  \end{bmatrix}
  \begin{bmatrix}x_{2k}\\x_{2k+1}\end{bmatrix}.
  \label{eq:rope}
\end{equation}
Here $\omega_k=10{,}000^{-2k/d_h}$ for $k=0,\ldots,d_h/2-1$;
$\operatorname{RoPE}$ applies these rotations columnwise at the supplied positions.
Thus all streams share one time axis, with context before and forecast tokens
after the origin.

\subsection{Condition Encoder}
History tokens remain fixed, $h^{(\ell)}=h^{(0)}$.
Nonfinal layers update recent and future tokens; the final layer
updates only future tokens. Independent attention and feed-forward norms are indexed by
$a$ and $f$. At layer $\ell=0,\ldots,3$, the projections are
\begin{equation}
  \begin{aligned}
    q_s^{(\ell)}&=\operatorname{RoPE}\!\left(
      \operatorname{RMS}_{q,s}^{(\ell)}(W_{q,s}^{(\ell)}\operatorname{LN}_{a,s}^{(\ell)}(s^{(\ell)})),\mathrm{pos}^s\right),\\
    k_s^{(\ell)}&=\operatorname{RoPE}\!\left(
      \operatorname{RMS}_{k,s}^{(\ell)}(W_{k,s}^{(\ell)}\operatorname{LN}_{a,s}^{(\ell)}(s^{(\ell)})),\mathrm{pos}^s\right),\\
    v_s^{(\ell)}&=W_{v,s}^{(\ell)}\operatorname{LN}_{a,s}^{(\ell)}(s^{(\ell)}).
  \end{aligned}
  \label{eq:encoder-qkv}
\end{equation}
Here $s\in\{h,r,f\}$ labels the stream and $\mathrm{pos}^s$ collects its token positions.
$q_s^{(\ell)}$ is defined for updated streams, while $k_s^{(\ell)}$ and
$v_s^{(\ell)}$ are defined for all three streams. $\operatorname{RMS}$ denotes
per-head normalization with learned scaling. Each updated stream jointly attends to all three
key-value banks:
\begin{equation}
  o_s^{(\ell)}=[v_{h}^{(\ell)},v_{r}^{(\ell)},v_{f}^{(\ell)}]\operatorname{softmax}\!\left(
    \frac{[k_{h}^{(\ell)},k_{r}^{(\ell)},k_{f}^{(\ell)}]^\top q_s^{(\ell)}}{\sqrt{d_h}}+M_s
  \right).
  \label{eq:encoder-joint-attention}
\end{equation}
Here $[\,,\,]$ concatenates token columns, softmax normalizes each column over keys,
and $M_s$ masks unavailable history or recent keys. Attention equations are written per head; head outputs
are concatenated before the output projection. Each updated stream uses its
own pre-normalized residual updates:
\begin{equation}
  \bar s^{(\ell)}=s^{(\ell)}+W_{o,s}^{(\ell)}o_s^{(\ell)},\qquad
  s^{(\ell+1)}=\bar s^{(\ell)}+\operatorname{FFN}_s^{(\ell)}(
  \operatorname{LN}_{f,s}^{(\ell)}(\bar s^{(\ell)})),
  \label{eq:encoder-update}
\end{equation}
where
$\operatorname{FFN}_s^{(\ell)}(x)=W_{2,s}^{(\ell)}[W_{u,s}^{(\ell)}x\odot
\operatorname{SiLU}(W_{g,s}^{(\ell)}x)]$,
with $\odot$ denoting elementwise multiplication. The encoder returns
$c=\operatorname{LN}_{\mathrm{out}}(f^{(4)})$.

\subsection{Flow Decoder and adaLN}
The noisy trajectory is patched as $x^{(0)}=\operatorname{Patch}_y(\tilde y_\tau)$. A Fourier
flow-time embedding and multilayer perceptron (MLP) produce
\begin{equation}
  e_\tau=\operatorname{MLP}([\sin(\tau\nu);\cos(\tau\nu)]),\qquad
  (\beta_{a},\gamma_{a},\alpha_{a},\beta_{f},\gamma_{f},\alpha_{f})
  =W_{\mathrm{mod}}\operatorname{SiLU}(e_\tau),
  \label{eq:adaln-parameters}
\end{equation}
where $\nu$ is a fixed frequency vector. The parameters $\beta,\gamma,\alpha$
provide feature-wise shift, scale, and residual gate, with $a,f$ indexing
attention and feed-forward branches. Each layer has its own modulation map
(indices omitted); plain LN has no learned affine parameters. For layer $\ell=0,1$, let
\begin{equation}
  \widehat x^{(\ell)}
  =\operatorname{LN}(x^{(\ell)})\odot(1+\gamma_{a})+\beta_{a}.
  \label{eq:adaln-attention}
\end{equation}
The projections $q_{x}^{(\ell)},k_{x}^{(\ell)},v_{x}^{(\ell)}$ follow Eq.~\eqref{eq:encoder-qkv},
with $s=x$ and $\widehat x^{(\ell)}$ replacing $\operatorname{LN}_{a,s}^{(\ell)}(s^{(\ell)})$.
Both future and condition tokens use $\mathrm{pos}^f$. The projected
condition $c_{d}=W_{c}c+b_{c}$ supplies per-layer cached keys and values
$(k_{c}^{(\ell)},v_{c}^{(\ell)})=\operatorname{KV}^{(\ell)}(\operatorname{LN}_{c}^{(\ell)}(c_{d}))$.
Here $\operatorname{KV}^{(\ell)}$ includes the key/value projections, followed by
$\operatorname{RMS}_{k,c}^{(\ell)}$ and RoPE on the keys. Decoder attention is one joint
softmax,
\begin{equation}
  o_{x}^{(\ell)}=[v_{c}^{(\ell)},v_{x}^{(\ell)}]\operatorname{softmax}\!\left(
    \frac{[k_{c}^{(\ell)},k_{x}^{(\ell)}]^\top q_{x}^{(\ell)}}{\sqrt{d_h}}
  \right),
  \label{eq:decoder-joint-attention}
\end{equation}
followed by gated residual updates
\begin{equation}
  \begin{aligned}
    \bar x^{(\ell)}&=x^{(\ell)}+\alpha_{a}\odot\bigl(W_{o,x}^{(\ell)}o_{x}^{(\ell)}\bigr),\\
    x^{(\ell+1)}&=\bar x^{(\ell)}+\alpha_{f}\odot\operatorname{FFN}_{x}^{(\ell)}\!\left(
      \operatorname{LN}(\bar x^{(\ell)})\odot(1+\gamma_{f})+\beta_{f}\right).
  \end{aligned}
  \label{eq:decoder-update}
\end{equation}
The output head uses its own flow-time-conditioned shift and scale,
\begin{equation}
  v_\theta(\tilde y_\tau,\tau,c)=\operatorname{Unpatch}\!\left(
  W_{v}[\operatorname{LN}(x^{(2)})\odot(1+\gamma_{o})+\beta_{o}]\right),
  \label{eq:decoder-output}
\end{equation}
where $(\beta_{o},\gamma_{o})=W_{o}\operatorname{SiLU}(e_\tau)$ and
$\operatorname{Unpatch}$ reshapes each hourly output into four quarter-hour
velocities. The condition key-value cache is fixed across ODE solver steps and
Monte Carlo samples; only the future-token projections are recomputed.

\subsection{Model and Training Configuration}

\begin{table}[H]
  \caption{Configuration of the reported \method{} model. SD: standard deviation;
  LR: learning rate; EMA: exponential moving average; MC: Monte Carlo.}
  \label{tab:appendix-model-config}
  \centering
  \footnotesize
  \setlength{\tabcolsep}{4pt}
  \begin{tabular}{ll@{\qquad}ll}
    \toprule
    Setting & Value & Setting & Value \\
    \midrule
    Parameters & 183.46M & Tokens (history / recent / future) & 336 / 24 / 168 \\
    Encoder / decoder blocks & 4 / 2 & Width / heads & 1,024 / 16 \\
    Patch size & 4 & Feed-forward expansion & 4$\times$ \\
    RoPE base & $10^4$ & Flow-time scale & $10^3$ \\
    adaLN initialization SD & $10^{-3}$ & Auxiliary-loss weight & $10^{-3}$ \\
    Optimizer & Muon + AdamW & Batch / maximum updates & 512 / 25,000 \\
    LR / warmup / schedule & $10^{-4}$ / 500 / constant & EMA decay / start & 0.999 / 1,000 \\
    ODE solver / steps & Euler / 20 & MC samples & 64 \\
    Sampling noise & Scrambled Sobol Gaussian & Checkpoint selection & Validation MSE \\
    \bottomrule
  \end{tabular}
\end{table}

During analytic decoding, the magnitude coordinate selects valid decimal
positions before inversion of the reflected digits. The auxiliary head
$f_\theta$ is linear and uses $\lambda_{\mathrm{aux}}=10^{-3}$ in
Eq.~\eqref{eq:fm-loss}.

All linear maps are initialized with Xavier-uniform weights and zero biases,
except the flow-time MLP, whose linear weights use a normal distribution with
standard deviation 0.02. Adaptive-normalization modulation weights in the flow
decoder use a zero-mean Gaussian with standard deviation $10^{-3}$. The final
modulation and output projection are initialized to zero.

We optimize two-dimensional hidden matrices in the transformer blocks with
Muon and all remaining parameters with AdamW. Training uses bfloat16 mixed precision. An
exponential moving average begins at update 1,000. Validation is run every 500
updates. Probabilistic and Michigan metrics are evaluated after model selection.

\subsection{Baseline Configuration}

The supervised baselines use the same county/origin sampler and available
covariates. DeepAR is a two-layer autoregressive LSTM with a Student-$t$
likelihood; TFT uses one LSTM layer, hidden size 64, four attention heads, and
nine quantiles from 0.1 to 0.9. CSDI treats the future outage window as missing
and retains four quarter-hour phases in each hourly patch. TimesFM~2.5\footnote{\url{https://huggingface.co/google/timesfm-2.5-200m-pytorch}} is
jointly fine-tuned with one shared linear covariate head, and Chronos-2 is fully
fine-tuned through its released interface. All checkpoints
are selected on validation data and fixed before test and Michigan evaluation. All
supervised baselines use AdamW with 500 warmup updates followed by a constant
learning rate. Chronos-2 uses the released interface's default learning rate of
$10^{-6}$. For the remaining baselines, we select the learning rate from
$\{10^{-4},10^{-3}\}$ using validation MSE.

\section{Metric Definitions}

\subsection{National Metrics}
\label{app:national-metrics}

Final generative evaluation uses $M=64$ trajectories generated from scrambled
Sobol Gaussian noise with 20 Euler steps. Let $N$ be the number of cases in the evaluated subset,
$T=672$, and $s(y)=\log_{10}(1+y)$. For sample models,
$\widehat{s}_{n,t}=M^{-1}\sum_m s(y^{(m)}_{n,t})$; quantile-only models use
their median. The case-equal mean squared error is
\begin{equation}
  \operatorname{MSE}
  =\frac{1}{NT}\sum_{n=1}^{N}\sum_{t=0}^{T-1}
  \left[\widehat{s}_{n,t}-s(y_{n,t})\right]^2.
  \label{eq:appendix-mse}
\end{equation}

For $\mathcal{Q}=\{0.1,0.2,\ldots,0.9\}$, let $\widehat{s}^{(q)}_{n,t}$
denote the predictive $q$-quantile of $s(y_{n,t})$. Weighted quantile loss is
\begin{equation}
  \operatorname{WQL}
  =\frac{2\sum_n\sum_t\sum_{q\in\mathcal{Q}}
  \rho_q\!\left(s(y_{n,t})-\widehat{s}^{(q)}_{n,t}\right)}
  {|\mathcal{Q}|\sum_n\sum_t |s(y_{n,t})|},\quad
  \rho_q(e)=\max\{qe,(q-1)e\}.
  \label{eq:appendix-wql}
\end{equation}
Sample-based models use empirical quantiles of the same 64 trajectories.

For interval level $1-\alpha$, let $l_{n,t}$ and $u_{n,t}$ be the empirical
$\alpha/2$ and $1-\alpha/2$ quantiles in transformed space. We report
\begin{align}
  \operatorname{Coverage}_{1-\alpha}
  &=\frac{1}{NT}\sum_{n=1}^{N}\sum_{t=0}^{T-1}
  \mathbf{1}\{l_{n,t}\le s(y_{n,t})\le u_{n,t}\},\\
  \operatorname{Width}_{1-\alpha}
  &=\frac{1}{NT}\sum_{n=1}^{N}\sum_{t=0}^{T-1}
  (u_{n,t}-l_{n,t}),
  \label{eq:appendix-interval}
\end{align}
with $\alpha=0.10$ in Table~\ref{tab:simulation-transfer}(a).

Temporal dependence is measured by the variogram score of order $0.5$
\citep{scheuerer2015variogram}:
\begin{equation}
  \operatorname{VS}^{(n)}
  =\frac{2}{T(T-1)}\sum_{t<t'}\Bigl(
  |s(y_{n,t})-s(y_{n,t'})|^{1/2}
  -\frac{1}{M}\sum_{m=1}^{M}
  |s(y^{(m)}_{n,t})-s(y^{(m)}_{n,t'})|^{1/2}\Bigr)^2,
  \label{eq:appendix-vs}
\end{equation}
and the reported value is
$N^{-1}\sum_{n=1}^{N}\operatorname{VS}^{(n)}$.

\subsection{Michigan Metrics}
\label{app:michigan-metrics}

The transfer evaluation uses all held-out Michigan counties, indexed by
$i=1,\ldots,C$, and the official 48-hour target. Michigan metrics are computed
in the original count space, whereas the national metrics above use $s(y)$.
\method{} uses weather over the 48-hour forecast horizon, following our national
evaluation protocol, while the reported competition entries follow their
original challenge input protocols.
Let $\widehat y_{i,t}$ denote the county-hour point forecast, with $t$ indexing
hours. The county-macro 24-hour RMSE is
\begin{equation}
  \operatorname{RMSE}_{24}=\frac{1}{C}\sum_{i=1}^{C}
  \sqrt{\frac{1}{24}\sum_{t=0}^{23}(\widehat y_{i,t}-y_{i,t})^2}.
  \label{eq:appendix-county-rmse}
\end{equation}
Define $\operatorname{RMSE}_{48}$ analogously over 48 hours. The DMDA average RMSE is
\begin{equation}
  \operatorname{RMSE}_{\mathrm{avg}}
  =\tfrac12\left(\operatorname{RMSE}_{24}+\operatorname{RMSE}_{48}\right).
  \label{eq:appendix-informs}
\end{equation}

For the IISE challenge metrics, $\kappa_i$ is county $i$'s 95th-percentile
training outage count. Let $\mathcal{A}_i^{\mathrm{normal}}=\{t:y_{i,t}<\kappa_i\}$ and
$\mathcal{A}_i^{\mathrm{tail}}=\{t:y_{i,t}\ge\kappa_i\}$ over the 48-hour target. Then
\begin{equation}
  \operatorname{RMSE}_{r}=\frac{1}{C}\sum_{i=1}^{C}
  \sqrt{\frac{1}{\max\{1,|\mathcal{A}_i^r|\}}\sum_{t\in\mathcal{A}_i^r}
  (\widehat y_{i,t}-y_{i,t})^2},\quad
  r\in\{\mathrm{normal},\mathrm{tail}\}.
  \label{eq:appendix-regime-rmse}
\end{equation}

Nonzero F1 labels a county-hour positive when its count is at least one. After
pooling all county-hours,
\begin{equation}
  F_1=\frac{2\mathrm{TP}}{2\mathrm{TP}+\mathrm{FP}+\mathrm{FN}}.
  \label{eq:appendix-f1}
\end{equation}
Here $\mathrm{TP}$, $\mathrm{FP}$, and $\mathrm{FN}$ count true positives,
false positives, and false negatives, respectively.

The 95\% interval uses the empirical 2.5th and 97.5th percentiles of the 64
generated trajectories. For lower and upper bounds $(l,u)$, its Winkler
score is
\begin{equation}
  W_{0.05}(y,l,u)=(u-l)
  +40(l-y)\mathbf{1}\{y<l\}
  +40(y-u)\mathbf{1}\{y>u\},
  \label{eq:appendix-winkler}
\end{equation}
averaged over all $C\times48$ county-hours.

For Table~\ref{tab:simulation-transfer}(b), the point summary for each metric
is selected on validation data and fixed before Michigan evaluation: the
median for normal RMSE, and the 80th percentile for average RMSE, tail RMSE,
and nonzero F1. Winkler uses the sample-based 95\% interval.

\clearpage
\section{Additional Forecast Examples}
\label{app:forecast-examples}

\begin{figure}[H]
  \centering
  \includegraphics[width=\linewidth]{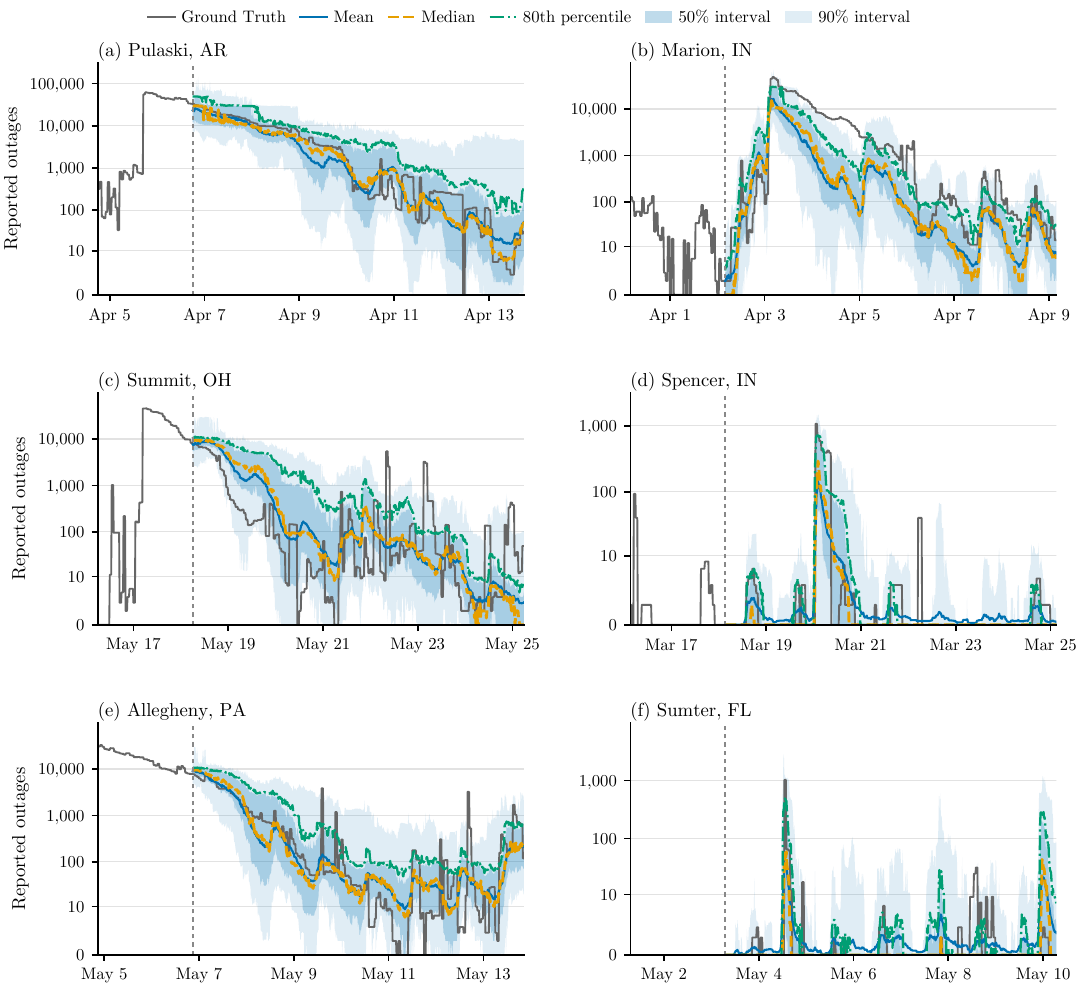}
  \caption{Additional recovery (left) and onset (right) forecasts. All dates are in 2025.}
  \label{fig:additional-forecasts}
\end{figure}

\end{document}